\pdfoutput=1

\documentclass[11pt]{article}
 \pdfoutput=1

\usepackage{acl}

\usepackage{times}
\usepackage{latexsym}

\usepackage[T1]{fontenc}
\usepackage[utf8]{inputenc}

\usepackage{graphicx}
\usepackage[]{xcolor}
\usepackage[linesnumbered,ruled]{algorithm2e}
\usepackage{amsmath,amsfonts,amssymb,bm}
\usepackage{booktabs}
\usepackage{bbold}

\usepackage{microtype}
\usepackage{soul}
\usepackage{enumitem}

\newcommand{\myul}[2][red]{\setulcolor{#1}\ul{#2}\setulcolor{red}}

\title{RSTGen: Imbuing Fine-Grained Interpretable Control \\into Long-Form Text Generators}

\author{
  Rilwan A.~Adewoyin\textsuperscript{1,4}, ~~Ritabrata ~Dutta\textsuperscript{2}, ~~Yulan ~He\textsuperscript{1,3}\\
\textsuperscript{1}Department of Computer Science, University of Warwick, UK\\
\textsuperscript{2}Department of Statistics, University of Warwick, UK\\
\textsuperscript{3}The Alan Turing Institute, UK\\
\textsuperscript{4}Department of Computer Science and Engineering, \\ Southern University of Science and Technology, China\\
  \texttt{\{rilwan.adewoyin,ritabrata.dutta,yulan.he\}@warwick.ac.uk}
}

\begin{document}
\maketitle

\begin {abstract}
In this paper, we study the task of improving the cohesion and coherence of long-form text generated by language models. 
To this end, we propose RSTGen, a framework that utilises Rhetorical Structure Theory (RST), a classical language theory, to control the discourse structure, semantics and topics of generated text. Firstly, we demonstrate our model's ability to control structural discourse and semantic features of generated text in open generation evaluation. Then we experiment on the two challenging long-form text tasks of argument generation and story generation. Evaluation using automated metrics and a metric with high correlation to human evaluation, shows that our model performs competitively against existing models, while offering significantly more controls over generated text than alternative methods.
\end {abstract}

\section{Introduction}

Controllable text generation has attracted much attention in recent years. Generating coherent and cohesive long-form text with controllable attributes is particularly challenging due to the relatively complex syntactic and semantic structures involved compared to short-text generation. Long-form text generation can find applications in automatic argumentation, motivational speech, and opinionated writing, to name a few.

A popular approach to controllable text generation is to design prompts, which control high-level linguistic features (i.e., text style and sentiment) by prefixing simple context to the input of a language model. These non-granular methods are often too coarse to allow different parts of the text to have diverse or contrasting features. Furthermore, they tend to focus on a single linguistic feature of text, meaning that extra frameworks such as PlugandPlay \citep{dathathri2019plug} are required to control multiple linguistic features.

To achieve improved content control, researchers have coupled a content planning module with a text generation module \citep{hua2020pair,hua2021dyploc}, in which the content planning module firstly generates a set of keyphrases and their corresponding positions in sentences and the generation module fills in the gaps.
To control syntactic structures of text, recent works have proposed a neuro-symbolic approach leveraging Dependency Structure Theory (DST) \citep{Shen2019OrderedNI,Du2020ExploitingSS}. 
Enforcing specific structures over the generated text has been proven to be useful in increasing coherence and cohesion of short-form text. However, a relatively large amount of DST information is required to encode comparatively short texts. Thus, they are unsuitable for use in long-form text generation tasks.

Similar to DST, Rhetorical Structure Theory (RST) \citep{MANNTHOMPSON+1988+243+281} is a classical tree-based interpretation of natural language. Whereas DST is most useful for intra-sentence interpretation, RST is most useful for inter-sentence interpretation of language. We propose a neurosymbolic framework that can imbue an RST understanding of text to existing language models. More  specifically, our framework (1) allows more fine-grained control of syntax, semantics and text structure; (2) utilises a well-defined rhetorical structure of language, thus offering better interpretability; (3) can be directly integrated into existing pre-trained language models such as GPT-2 \citep{radford2019language} and BART \citep{lewis2019bart}.

We evaluate our proposed framework on two tasks: argument generation and story generation. We show that our proposed framework improves upon existing content control approaches on automatic evaluation metrics including BLEU \citep{papineni-etal-2002-bleu}, METEOR \citep{denkowski-lavie-2014-meteor} and generates better text in terms of grammar and coherence measure. We further demonstrate
our model's ability to generate text with control over syntactic, semantic and discourse features. 
Our main contributions can be summarised below:
\begin{itemize}
    \item We propose a novel framework to imbue the RST information into pre-trained language models.
    \item We develop the first neurosymbolic framework that provides interpretable fine-control over syntax, semantics, discourse structure, topical keywords and keyword positions.
    \item We demonstrate the superiority of our proposed framework over existing planning and control methods for two long-form text generation tasks.
\end{itemize}

\section{Related Work}

We discuss several lines of research in controllable text generation.

\paragraph{Prompts}
For prompt-based control, a \textit{context} is prepended to a language model input. The prepended item is related to the type of desired output.
This method has been used to manipulate the syntax \citep{duvsek2016sequence, goyal2020neural} and semantics \citep{wen2015semantically,chen2019multi} of the output. Alternatively, 
the prepended item can provide semantic control in order to cover a given topic \citep{wang2019topic}, mention specified entities \citep{fan2019strategies}, display a certain attribute \citep{pmlr-v70-hu17e, balakrishnan-etal-2019-constrained}, or even exhibit a style of text \citep{keskar2019ctrl}. These methods suffer from the inability to exert fine-control, that is, a change in any one of the input prompts will change the whole generated text. Furthermore, all these works 
utilise non-expansive features for their prompts, which prevents them from making iterative improvements to existing generated text.


\paragraph{Content Planning}
Despite the impressive progress made in many generation tasks, earlier neural models are known to produce low-quality content
\cite{wiseman2017challenges}, often with low relevance, and poor discourse structure \citep{zhao2019rethinking}. Content-based planning approaches were added into neural systems to enhance content relevance \citep{moryossef2019step,yao2019plan,hua2019sentence}. \citet{hua2021dyploc} extended previous content planning approaches by dividing the content plan into different types of information; `Entities', `Claim' and `Concepts'. An alternative planning approach focuses on where the key phrases should be placed in the generated text in order to improve coherence \citep{hua2020pair}. The text generator is then conditioned on the provided key phrases and their respective positions in text.
Similar to our approach, these methods allow users varying levels of custom control by manual augmentation of the planning module output. 

\paragraph{Syntactic or Discourse Control}
Various works utilised syntactic parse trees with the transformer structure to gain syntactic control or improve interpretability on short-form text generation tasks \citep{li2020transformerbased,nguyen2020treestructured}. With a focus on improving long-form text generation, \citet{ji2021discodvt} used a Variational Autoencoder (VAE) structure to model RST discourse relations between successive elementary discourse units (spans of text). 

We build upon their approach by using a more expressive binary tree formalisation of RST. This formalisation extends the modelling of sequential elementary discourse units by also modelling nuclearity and relationship between non sequential discourse units.

Extending on previous works, \citet{wang-etal-2019-tree} attempted to couple tree structures and transformers. We instead embed the tree structure of RST into transformers through constrained attention in a separate RST embedding dimension.


\section{Rhetorical Structure Theory}
\begin{figure}[htbp]
    \centering
    \includegraphics[width=\linewidth]{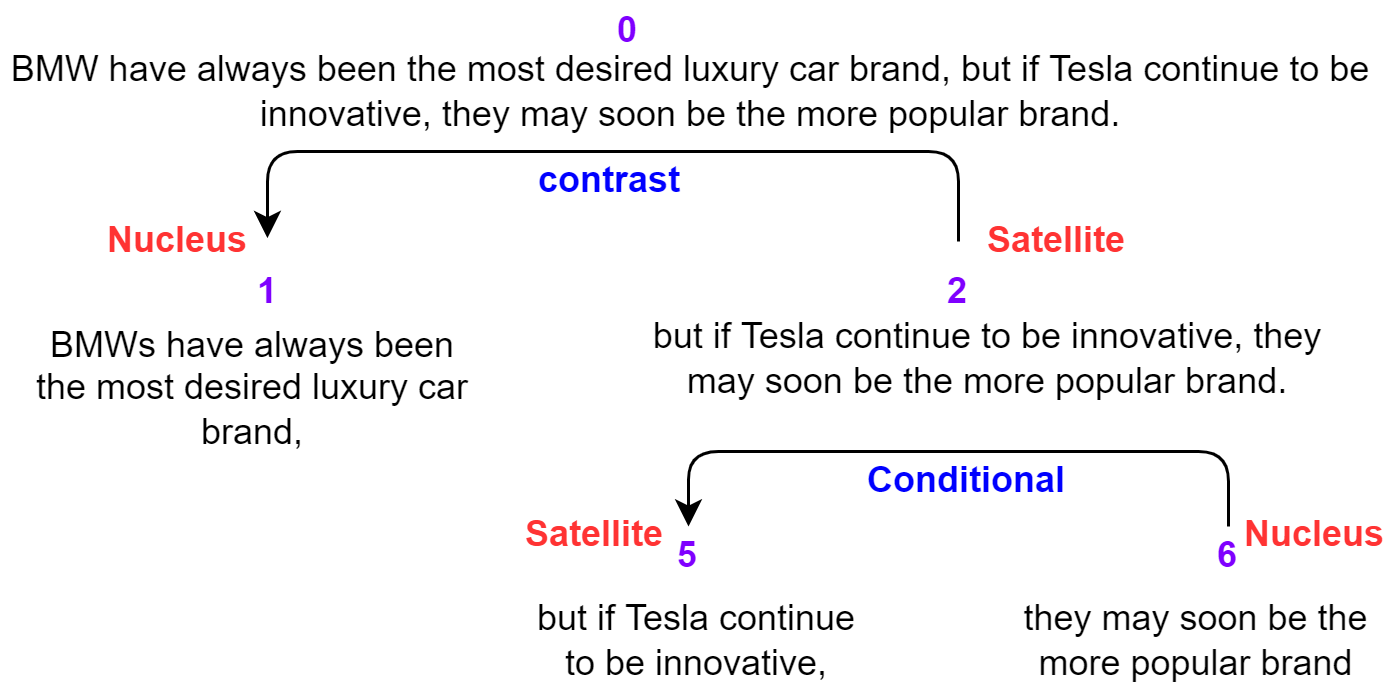}
    \caption{\textbf{RST example: } An example text (top) interpreted through RST schema. The text is broken down into a binary tree with zero-indexed \textcolor{purple}{numbered node positions}. \textcolor{blue}{Relations} exist between the two sibling nodes of a parent node. Each pair of sibling nodes, have a coupled \textcolor{red}{nuclearity} label. Nodes with no children, are Elementary Discourse Units (EDUs).}
    \label{fig:RST_diagram}
\end{figure}

Rhetorical Structure Theory (RST) provides a formal structure for interpreting language based on relations between parts of text. Each structure is defined by three sets of features: the binary tree structuring of Discourse Units (EDU)s, the nuclearity between sibling DUs, and the relationship between sibling DUs. An example RST tree using the schema provided by \citep{MANNTHOMPSON+1988+243+281} as shown in Figure~\ref{fig:RST_diagram}.

\paragraph{Binary Tree Structuring of Discourse Units}
To form a binary discourse tree $R$ from text $t$, the text must be successively divided into smaller subsets of text. In this manner, node 0 represents the full text, with subsequent nodes representing sub-texts, as in Figure~\ref{fig:RST_diagram}. The text at each node is called a Discourse Unit, while the nodes at the leaves of the tree are Elementary Discourse Units (EDUs).

\paragraph{Sibling Nodes $-$ Relations: }
A parent node is a non-terminal node which has two child nodes, referred to as siblings. Each parent nodes' children, has a RST relation $r_{i}, i\in \mathbb{N}_{<20}$ describing the syntactic purpose of each child node relative to the other. In Figure~\ref{fig:RST_diagram}, the RST relations are presented in blue texts.\footnote{Table~\ref{tab:RST_Relations} in the Appendix details the 19 possible RST Relations.}

\paragraph{Sibling Nodes $-$ Nuclearity: }
Associated with each pair of sibling node's relation is a joint nuclearity labelling, $n_{j}, {j\in \mathbb{N}_{<5}}$. This nuclearity explains the role of each sibling node's discourse unit relative to the other sibling node. Each joint nuclearity labelling between pair of sibling nodes must include at least one Nuclei.\footnote{Table~\ref{tab:RST_Nuclearity} in the Appendix details the 4 possible RST Nuclearity values.}

Formally, we define a binary RST Tree $R$ as a collection of its parent nodes $\{ \bm{v}^{r,n}_{l} \}$, where $\bm{v}^{r,n}_{l}$ is a parent node at position $l$, with a RST relation $r$ and a Nuclearity $n$ between its two children. For example, $v^{\text{Conditional},SN}_{2}$ represents node 2 in Figure~\ref{fig:RST_diagram}.
We refer to the group of relations in an RST Tree, $R$, as $R^r$. Similarly, the group of Nuclearity values is referred to as $R^n$.

\section{RSTGen: RST-Dependent Long-form Text Generator}

\begin{figure}[htbp]
    \centering
    \includegraphics[width=1.0\linewidth]{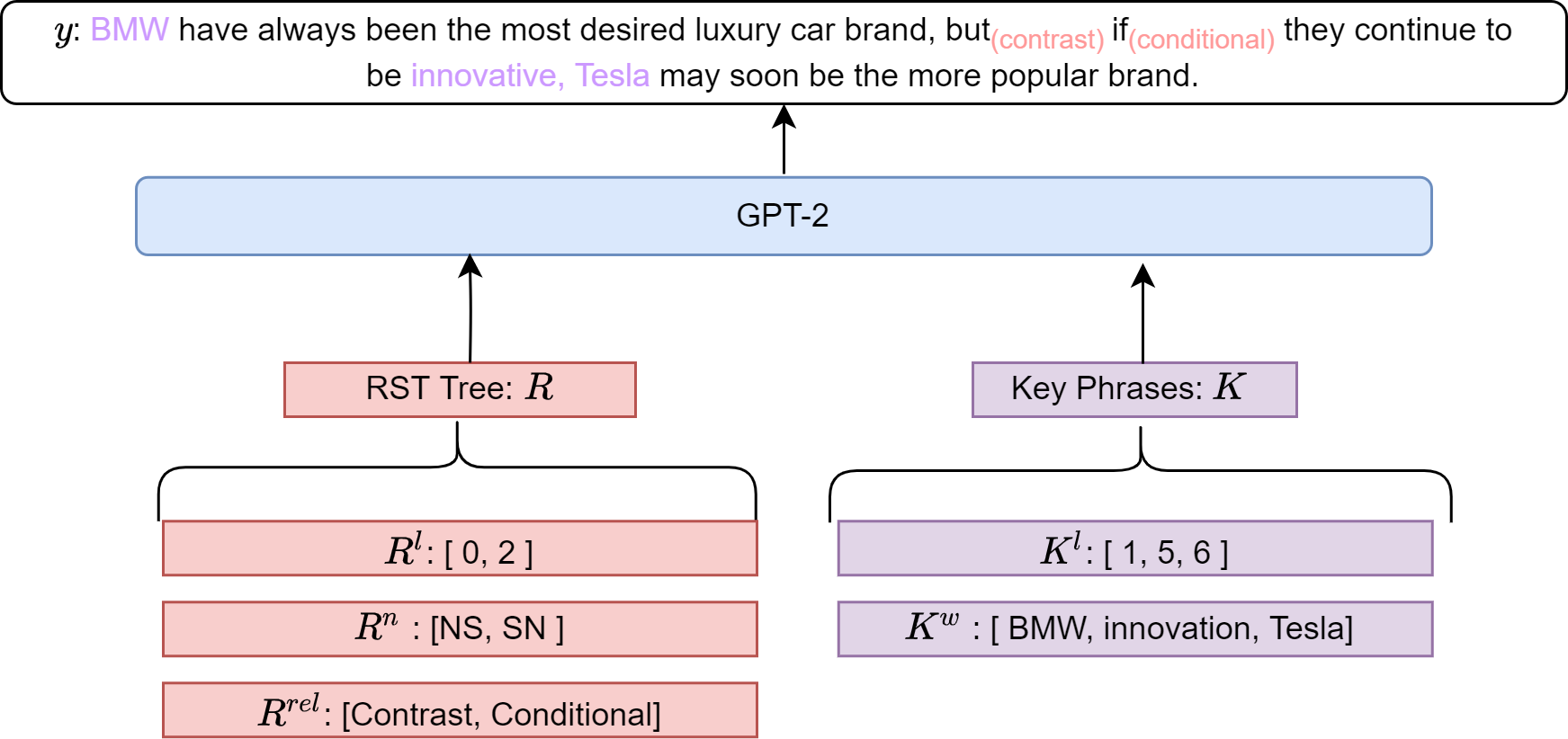}
    \caption{\textbf{ RSTGen Structure: } Our RST controlled text generation framework can be built upon any pre-trained language model. We input RST tree structure information as $R$, comprising the Nuclearities $R^n$, RST Relations $R^r$ and RST node positions $R^l$, and key phrase information at EDUs as $K$, comprising the key phrases $K^w$ and the RST node positions of the key phrases, $K^l$. The discourse structure of the output text $y$ is controlled by $R^n$ and $R^l$. The syntax of $y$ is controlled by $R^n$, $R^l$ and $K^l$. The semantics is controlled by $R^r$ and $K^w$. Our framework creates a prefix-embedding from the encodings for $R$ and $K$, after which text is generated as a continuation.}
    \label{fig:RSTGen_model}
\end{figure}

We propose a long-form text generation approach, call RSTGen, which provides control over semantics and structure of text, with the aim of improving \emph{coherence} and \emph{cohesion} of the generated text. 
\emph{Coherence} is concerned with how well constituent sentences/chunks follow on from the previous sentence/chunk. This can be interpreted as the topic of each subsequent sentence/chunk being relevant to the previous sentence/chunk \citep{10.2307/3586466, waller_2015}. \emph{Cohesion} describes the way in which text is tied together by linguistic devices such as \textit{Therefore..., However..., In addition to} \citep{waller_2015}. This can be interpreted as how smooth the model transitions between the types of sentences. With an RST interpretation, \emph{cohesion} is loosely related to the RST relations between different nodes in the binary tree, while \emph{coherence} is related to the RST relations and key phrases at EDUs.

\paragraph{Task formulation} Our input consists of (1) an encoding for a binary RST Tree $R$, (2) a set of keyphrases information $K$ that are relevant to the text. The RST encoding is formed as sets of three pieces of information for each parent node in the binary RST Tree. These include Nuclearity $R^n$, RST Relation $R^r$ and RST node position $R^l$. The keyphrase information contains two pieces of information. These are $K^w$ and $K^l$, the key phrases and the RST node positions of the key phrases. The model has been trained to work with varying levels of specification for the context information. 

\paragraph{Partial Context Provision}
In practice, we do not expect users to provide the full RST Tree or key phrases since often only a subset of features of the context information may be known or needed. For example, in Figure \ref{fig:RSTGen_model}, the goal is to produce a text contrasting the innovative ability of Microsoft and Tesla. A user only needs to provide partial information, e.g., the keywords, $K^w = [BMW, Tesla, innovation]$ and the RST relation, $R^r \supset ['Contrast']$. Our model will be able to automatically predict an RST tree following 
a conditional sampling method we have designed in Section \ref{sec:RSTPredictor}.

While a user will often want to specify the key phrases themselves, we could also employ automated methods to create an expanded set. For example, existing Planning Modules \citep{hua2020pair, hua2021dyploc} can be used to generate a set of keywords based on some prior information.

In what follows, we describe our proposed RSTGen in more details.

\subsection{Tokenisation}

While our RST controlled text generation framework can be built upon any pre-trained language model, we use GPT as an example here. The GPT Tokeniser is used to tokenise the target text $y$ and key phrases $K^w$ to a set of word tokens. 
Two new tokens are added to the tokeniser: `\texttt{<rst>}' and `\texttt{<kp>}'. The former is prepended to the RST tree $R$, and the latter is prepended to each key phrase.

\subsection{Embedding Layers}

Our framework requires three additional embedding layers to facilitate the encoding of the RST position, RST nuclearity and RST relation information presented in $R^l, R^n$ and $R^{r}$. 
These embedding layers are designed to produce vectors that match the size $v$ of hidden vectors in the base model.

We use a RST relation Embedding Matrix $W_r\in \mathbb{R}^{19\times 768}$ to encode $R^{r}$ as there are 18 possible RST relations and a pad token, and an RST Nuclearity Embedding matrix $W_n\in \mathbb{R}^{4\times 768}$ to encode $R^n$, which consists of 3 possible nuclearity labels and a pad token.

To embed RST node position encodings $R^l$, we create a novel embedding layer designed to capture the positional relationships of nodes in a binary tree, in a space efficient manner.
The intent is to explicitly capture the relationship between a node and its ancestor nodes. Our embedding layer features a non-trainable embedding matrix $W_{pe} \in \mathbb{R}^{\mathrm{max\_rst\_node} {\times} \mathrm{tree\_depth} }$, a trainable feed forward layer $W_{pff}\in \mathbb{R}^{ \mathrm{tree\_depth} {\times} 768 }$ and a Gelu activation layer $f_{g}(\cdot)$. In our experiments $\mathrm{max\_rst\_node}=\mathrm{4094}$ and $\mathrm{max\_tree\_depth}=\mathrm{12}$.

The $i$-th column in the non-trainable embedding matrix $W_{pe}$ is a vector representing the position of node $i$, in terms of a sequence of Lefts (L) and Rights (R) required to travel from the root node $0$ to node $i$. Left is encoded as $-0.05$ and Right as $0.05$. For example, the vector at node position $5$ in $W_{pe}$ in the example RST tree shown in Figure~\ref{fig:RST_diagram} is encoded as $[0.05, -0.05, 0, 0.... ]$ with the remaining $0$s representing padding values up to the length $\mathrm{max\_tree\_depth}$. 

It is important to note that in our RST Tree encoding $R$, a parent node $\bm{v}_{i}$ is labelled with the relation $r_{i}$ and nuclearity $n_{i}$ connecting its children. As such our encodings for a $R^r, R^l$ and $R^n$ do not include the leaf nodes with no children, but still represent $R$. This allows our encodings, for a Binary RST Tree with $N$ nodes to be sequences of length $\lceil N/2 \rceil$.

\subsection{RST Predictor}
\label{sec:RSTPredictor}

We use an RST Predictor to predict the relation and nuclearity of a child node conditional on the RST relation, nuclearity and position of their parent node.
For example we predict the left child node $\bm{v}^{r_i,n_j}_{2l+1}$, by modelling the following conditional distribution 
$\bm{v}^{r_i,n_j}_{2l+1}  \sim p\big( r, n | l_{\mathrm{parent}}, r_{\mathrm{parent}}, n_{\mathrm{parent}}, \mathbb{1}\small(\bm{v}_{2l+1}=\text{left child}\small)$. A full RST Tree can be predicted by iteratively repeating this one-step sampling. 



\begin{figure}[!htbp]
\centering
    \includegraphics[width=0.75\linewidth]{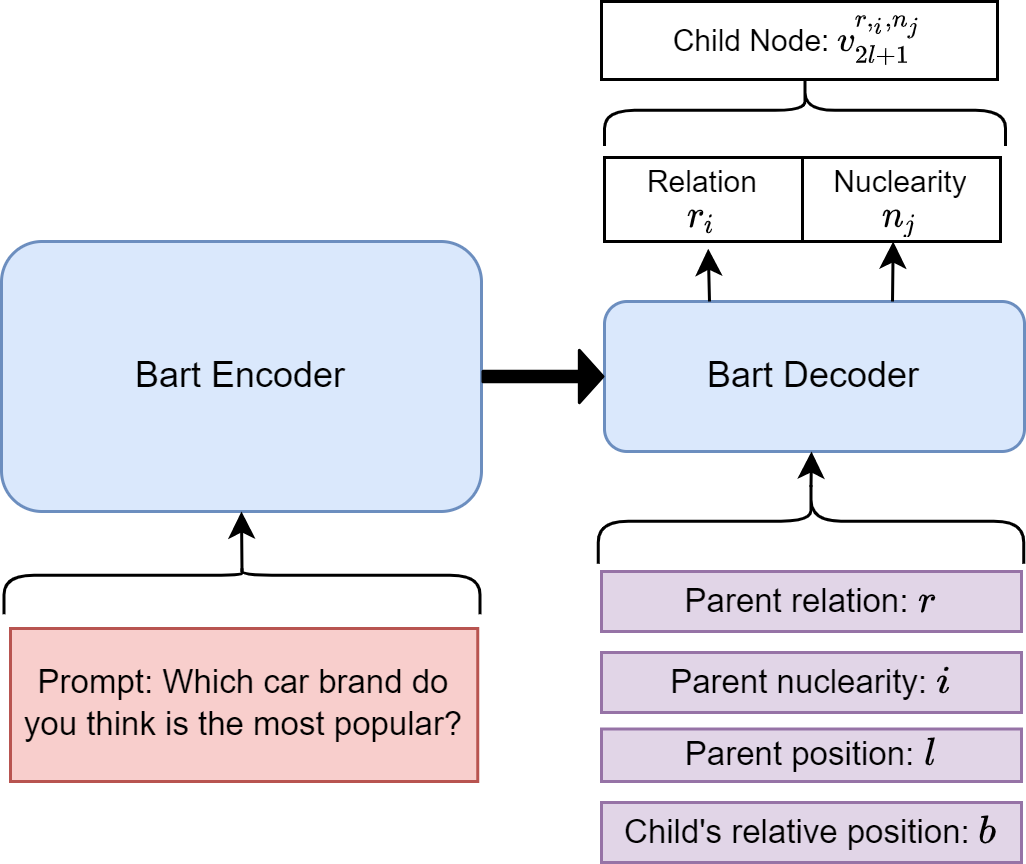}
\caption{\textbf{Neural RST Sampler:} Our Neural RST Sampler can be used to iteratively predict an RST Tree, conditioned on a prompt which is passed to the Encoder. The Decoder takes as input the summation of embeddings for the parent' nodes relation $r$, nuclearity $n$ and position $l$. Further we also pass in $b$ which represents whether the target child node is a left or right child.}
\label{fig:neural_RST_sampler}
\end{figure}

We 
propose a neural sampler method to estimate the conditional distribution $p(\mathord{\cdot}|\mathord{\cdot})$. It is trained on the RST Trees observed in the training set of the RST Annotated Dataset introduced in Section~\ref{sec:tasks_and_datasets}. Our neural RST Sampler, depicted in Figure~\ref{fig:neural_RST_sampler}, uses the BART \citep{lewis2019bart} model. The encoder takes a prompt as input. The decoder is re-initialised and reduced to two layers. The decoder takes four inputs: (1) The parent nodes' relation $r$; (2) The parent node's nuclearity $n$; (3) The parent node's RST position $l$; (4) A vector $b$ indicating whether the target node is a left child node or right child node. This vector $b$ is calculated as the difference between the node position embeddings for the parent node and the target child node encodings.

We use the Embedding Layers structures from our RSTGen model to encode the inputs (1-4) to the Encoder. The BART decoder contains two classification heads, to predict relation and nuclearity.

Two approaches can be used to guide our RST predictor to produce more relevant RST Trees. Firstly, the training set can be limited to a specific domain, such as argumentative text, to introduce bias the prediction towards a specific linguistic style. Secondly, during inference the predictive distribution can be augmented, to ensure specific structural properties are achieved. In our experiments, we use this to predict RST Trees with a specific size. We have extended this to other structural features such as specifying RST relation frequency, and position.

\subsection{RST-Aware Attention (R.A.A.)}
\label{sec:RST-attention}


To better encode the RST tree structure in language model training and to improve the coherence of text by reducing the occurrence of hallucination in long-form generated text, we propose a novel attention scheme, called RST-aware attention. In particular, to generate a word token at position $i$, we create dynamic attention masks $\bm{m}_i$ allowing the hidden state $h_i$ to focus only on structurally relevant hidden representations $h_j$, $j\ne i$. The hidden representation $h_j$ is structurally relevant to $h_i$ if $h_j$'s associated RST node is an ancestor of the RST node associated to $h_i$. The RST-aware attention is described in Algorithm~\ref{alg:RSTAttention}.

\begin{algorithm}[htb]\small
    \SetKwInOut{Input}{Input}
    \SetKwInOut{Output}{Output}
    
    \Input{ Hidden representations: $h_i$, $h_j$, $j\ne i$, and their associated RST nodes $\mbox{v}_l$, $\mbox{v}_m$ at positions $l, m$ respectively.    }
    \Output{ Updated hidden representation $h_i'$.} 
    \For{$j \ne i$} {
        \uIf{isParentNode$(\mbox{v}_l, \mbox{v}_m )$}{
            $m_{i,j} = 1$\
        }
        \uElse{
            $m_{i,j} = 0$\
        }
      }
    \Return $h_i'\leftarrow \mbox{MaskedAttention}(h_i, \{h_j\}_{j \ne i},\bm{m}_i )$
    \caption{RST-aware attention}\label{alg:RSTAttention}
\end{algorithm}

In the above process, we need to first detect the RST node position of the word token to be generated. 
We do this in a sequential manner with the help of an EDU splitter \citep{wang-etal-2018-toward}, which can detect EDU boundaries in text. At the start of text generation, we set the initial EDU node position as the leftmost leaf node in the RST tree and then proceed to generate the first token. Afterwards, we use the EDU splitter to detect whether an EDU boundary has occurred. If no boundary is detected, we continue with the generation of the next word token; otherwise, we infer the next EDU node position as the second leftmost leaf node in the RST tree. The above process is repeated until the `\texttt{end of sequence}' token is generated or until the final child EDU node is generated. In practice we use heuristics to avoid the need to perform an EDU boundary detection at each generation step.

\section{Open Generation Evaluation}

We first train a model using our RST annotated dataset described below. Then we analyse the model's ability to control semantic, syntactic and text structure.

\begin{table}[h!]\small
\centering
\begin{tabular}{@{}lc@{}}
\toprule
Subreddit                             & \% of Training Instances \\ \midrule
r/CasualConversation   & 41.7                         \\
r/changemyview         & 19.1                         \\
r/DebateReligion       & 15.8                         \\
r/PoliticalDiscussion  & 9.62                         \\
r/relationship\_advice & 7.94                         \\
Other Subreddits                      & 6.84     \\ \midrule
Total                                 & 965,411 samples       \\
\bottomrule
\end{tabular}
\caption{\textbf{Statistics of RST Annotated Dataset:} The majority of our dataset is sourced from 5 subreddits which exhibit a tendency to produce longer and more complex texts.}
\label{tab:RSTAnnotatedDatasetDetails}
\end{table}

\begin{table*}[htb]
\centering
\resizebox{2\columnwidth}{!}{%
\begin{tabular}{@{}lll@{}}
\toprule
\textbf{Input}&&\\
\midrule
Key phrases:     & ``\emph{good mouse alternative}" and ``\emph{none}"  \\
Text prompt: & ``\emph{It would be}"  \\
RST structure:           & \{Relation: $r1$, Nuclearity: SN, Node position: 0\}, \{Relation: $r2$, Nuclearity: NS, Node position: 2\}                 \\ 
\midrule
\textbf{Input Relations $(r1, r2)$}          & \textbf{Generated Text} \\ 
\midrule
Attribution,   \underline{Explanation} & It   would be a good mouse alternative, \textcolor{red}{because} none of us are going to be able to see. \\
\underline{Background},   Evaluation   & It   would be a good mouse alternative, \textcolor{red}{since} none of us have ever seen one.& \\
Background,   \underline{Condition}    & It   would be a good mouse alternative \textcolor{red}{if} none of them were good.                        & \\
\underline{Contrast}            & It   would be a good mouse alternative, \textcolor{red}{but} none of them are as good as a good one.       &\\ \bottomrule
\end{tabular}}
\caption{\textbf{Short Text Semantic control: }Examples of text generated by varying the RST relations. Words highlighted in red in the generated text indicate the corresponding underlined RST input relations. }
\label{tab:SemanticControlShort}
\end{table*}

\paragraph{RST Annotated Dataset}
We use the ConvoKit API \citep{chang2020convokit} to collect a total of over 965k text examples from Reddit to use as the training set. Table~\ref{tab:RSTAnnotatedDatasetDetails} provides a detail regarding the division of the dataset between different subreddits. The following subreddits comprise the majority of our dataset, \emph{DebateReligion}, \emph{RelationshipAdvice}, \emph{Politics} and \emph{ChangeMyView}. These subreddits contain many long texts with opinions or persuasive style of writing. Further, we choose large samples from the subreddit \emph{CasualConversation} to complement the limited range of language present in the former subreddits. Each sample contains a reddit post, its RST tree generated using the best statistical RST parser \citep{feng-hirst-2012-text} and keyphrases extracted from the post using the PageRank inspired algorithm, TextRank \cite{mihalcea-tarau-2004-textrank}. The keyphrase extraction process is detailed in Appendix \ref{apdx:text_rank}.

\paragraph{Semantic Control}
In Table~\ref{tab:SemanticControlShort} we show the generated short text conditional on various RST relations. For all four examples the same key phrases of ``\emph{a good mouse alternative}" and ``\emph{none of this}" were passed to the model. Furthermore, we provide the first three words ``\emph{It would be}" to the decoder as a text prompt. The text is generated by varying the RST relations, $r_1$ and $r_2$. We can observe that the generated text varies depending on the desired RST relations input to the model.

\paragraph{Structural Discourse Control}

\begin{figure}[!htbp]
\centering
    \includegraphics[width=\linewidth]{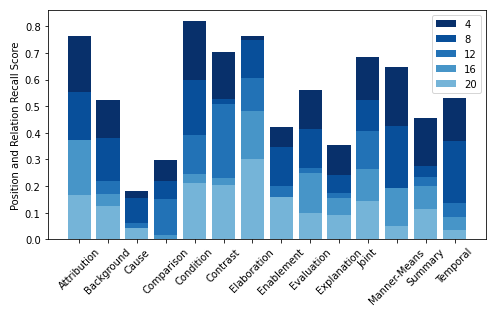}
\caption{\textbf{Structural Discourse Control:} RST relation and position recall scores calculated for each RST relation from the generated text with its lengths measured by the number of Elementary Discourse Units in the range of $\{4,8,12,16,20\}$.}
\label{fig:SyntaticControl}
\end{figure}

Here we use a reconstruction style test to evaluate the ability of our model to include the target relations in the generated text at the correct position. To do this, we allow our model to generate text $\hat{y}$ using the RST tree $T$ as the conditioning factor. We then use an RST parser to generate a RST tree $\hat{T}$ from the generated text $\hat{y}$. For each relation $r$, we calculate a recall score as the proportion of nodes with relation $r$ in $T$ that also occur in $\hat{T}$ at the same position $l$. The results are shown in Figure~\ref{fig:SyntaticControl}. We include results for different lengths of text, measured by elementary discourse units.

We observe that our model achieves strong levels of control for the following relations: \emph{Attribution, Background, Condition, Contrast, Elaboration, Joint, Manner-Means}, and \emph{Temporal}. We believe that the weakened performance on \emph{Cause} and \emph{Comparison} is due to their respective similarity to \emph{Attribution} and \emph{Contrast}. We omit topic change since our datasets contains texts mostly constrained to a single topic.

\begin{figure}[!htbp]
\centering
    \includegraphics[width=\linewidth]{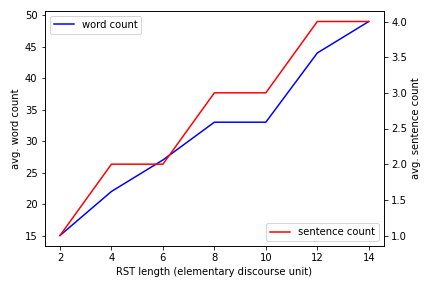}
  \caption{\textbf{Text Length Control:} RSTGen exhibits the ability to implicitly control the length of generated text by controlling the size of the RST structure provided as context. In this figure we present the average word count (left $y$-axis) and average sentence count (right $y$-axis) against the length of the RST encoding passed to RSTGen.
  }
\label{fig:text_length_control}
\end{figure}

\paragraph{Text Length Control}
By editing the length of the RST encoding, we gain strong control of the length of the generated text. 
Here we fix the key phrases and vary the length of the RST context (i.e., number of EDUs) from 2 to 14 to demonstrate the increasing length of generated text. Results are depicted in Figure~\ref{fig:text_length_control}.
We believe that our method provides a more natural way to control text length when compared to the heuristic methods of fine tuning text generation hyper-parameters such as repetition penalty, length penalty and minimum length. 

\begin{figure}[!htbp]
\centering
    \includegraphics[width=\linewidth]{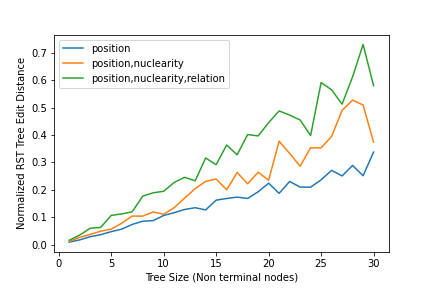}
  \caption{ \textbf{Normalised RST Tree Edit Distance: } This figure shows the similarity between the RST structure of RSTGen output text and the RST encoding passed as input to RSTGen. We extend Tree Edit Distance to RST Trees to calculate three metrics based on combinations of node position, node nuclearity label and node relation label: \underline{(1) Simple Structure:} Tree distance including only node positions; \underline{(2) Complex Structure:} Tree distance including node positions and nuclearity; \underline{(3) Complete Structure:} Tree distance including node position, nuclearity and relation. }
\label{fig:tree_edit_distance}
\end{figure}

\paragraph{RST Tree Edit Distance}
Here we extend the Tree Edit Distance to RST Trees to evaluate how well RSTGen is able generate text that adheres to the RST encoding passed to it. Specifically, we investigate how well this model performs as the RST structure becomes increasingly long. For this experiment, we pass an RST Encoding and a set of keywords to our model and generate a text $y$. We then extract an RST tree from the generated text $y$ using an RST parser. An edit includes the following operations: Changing/Removing the relation or nuclearity label of the node at position $l$ with cost 1; Changing the position of a node to it's sibling node with cost $1$; and Deleting/Inserting a blank node at position $l$ with cost 3. For a tree of size $s$ we use a normalising factor of $3 {\times}s$, the cost of creating the tree. Our normalising factor does allow distances over 1.
We observe from Figure \ref{fig:tree_edit_distance} that generated text is able to adhere to correct RST structure in terms of RST node positions relatively well. However, when considering nuclearity and relation, he inaccuracy of RST structure grows significantly.

\section{Evaluation on Argument Generation and Story Generation}
\label{sec:tasks_and_datasets}

We evaluate RSTGen for the tasks of argument generation and story generation. These tasks require our model to output coherent and cohesive text. 

\subsection{Experimental Setup}

We describe the datasets used for fine-tuning our model, baseline models and the methods used to ensure fair comparison. Training detail and hyperparameter setting can be found in Appendix~\ref{apdx:reproducibility}.

\paragraph{Datasets}

We use two different datasets for argument generation and story generation, respectively.

\noindent\underline{CMV Dataset}
We use the argument generation dataset first introduced in \citep{hua2021dyploc}. This dataset contains pairs of claims and counter-arguments extracted from titles and responses, respectively, from the Reddit \emph{ChangeMyView} (CMV) forum. This dataset uses posts dated in 2019, which prevents overlap with our training set for the RST Annotated Dataset. The goal of this task is to generate the counter-argument given the initial claim. 

\noindent\underline{Writing Prompts Dataset}
We use the Story generation dataset created by \citet{fan-etal-2018-hierarchical} and utilised in \citep{ji2021discodvt}. This dataset contains pairs of prompts and stories extracted from the \emph{WritingPrompts} subReddit. The prompt is an introduction to a story that must be extended. We limit each story to be up to 270 tokens.

\begin{table*}[tb!]
\centering
\resizebox{2\columnwidth}{!}{%
    \begin{tabular}{@{}llcccc@{}}
        \toprule
        Model           & Control Method        & Grammar $\uparrow$ & Redundancy $\uparrow$ & Focus $\uparrow$ & Structure and Coherence $\uparrow$\\ \midrule
        CTRL            & Text prompt            & \textbf{0.84}         & -0.16        & -0.12       &       -0.76                  \\
        PAIR            & Keywords $+$ positions        & 0.79         &   \textbf{0.00}     & \textbf{-0.02}      & -0.78 \\
        DYPLOC          & Categorised keywords      &  0.73       &     -0.10   & -0.03      &    -0.79                    \\ \midrule
        RSTGen & RST $+$ PAIR Planner   & 0.81   & -0.03           & -0.03   &        -0.77 \\
        RSTGen & RST $+$ DYPLOC Planner & 0.83    & -0.02           & -0.04   & \textbf{-0.75}       \\
        RSTGen & RST $_{predicted}+$ PAIR Planner   & 0.75   & -0.05 & -0.06   & -0.88                 \\
        RSTGen  & RST $_{predicted}+$ DYPLOC Planner & 0.76   & -0.06 & -0.05   & -0.86\\        \midrule
        RSTGen w/o R.A.A.  & RST $+$ PAIR Planner   & 0.71    & -0.04 & -0.08   &    -0.81\\
        RSTGen w/o R.A.A.  & RST $+$ DYPLOC Planner & 0.65    & -0.06  & -0.09   &    -0.80\\
        \bottomrule 
    \end{tabular} %
    }
    \caption{\textbf{Argument Generation Evaluation:} the performance of baseline models and our RST frameworks with different control methods. The shortform R.A.A refers to the RST-aware attention described in Section~\ref{sec:RST-attention}.}
    \label{tab:arg_gen_performance1}
\end{table*}

\paragraph{Baselines}

For argument generation, we evaluate the following baselines:

\noindent\underline{CTRL}. \citep{keskar2019ctrl} uses text prompts to control the style of output. To trigger the text style found in the r/changemyview subreddit,  we prepend the text `\texttt{Opinion Title:}` to the opening statement. This model is approximately 10 times larger than RSTGen and we assume that it achieves performance gains primarily through its larger size.

\noindent\underline{PAIR} (Planning and Iterative Refinement). \citet{hua2020pair} devised a two-step generation process. The first step uses a fine-tuned BERT to allocate a position to each pre-defined keyword, in order to create a template for the text to be generated. Then a fine-tuned BART model is used to fill the template.

\noindent\underline{DYPLOC} (Dynamic Planning of Content using Mixed Language Models). \citet{hua2021dyploc} proposed first categorising the keywords into concepts and entities. A BERT-based Planner is used to expand this set of concepts and entities, while another is used to generate the claim. The claim provides a brief summary of the extended answer. Four separate BART Encoders and a BART Decoder are used to convert the expanded set of concepts and entities into an argumentative text.

For both PAIR and DYPLOC, we use their gold standard plan as input to the model, this follows the headline results given in both papers. In pursuit of fair testing, we convert the gold standard plans of PAIR and DYPLOC to two RST encodings. More concretely, for PAIR, we convert its gold standard template, consisting of a set of phrases and their corresponding starting positions in the text to be generated, into a RST encoding. These phrases become the keyphrases for RSTGen. The RST node position for each key phrase can be determined by parsing the true RST structure of the text. For DYPLOC, first we collect the gold standard set of concepts/entities and claims, and filter out any repeated words phrases. Then, the RST node position for each word/phrase can be determined in a similar manner to PAIR.


For Writing Prompts, we compare our proposed approach with the following models:

\noindent\underline{DiscoDVT} \citep{ji2021discodvt}.
To the best of our knowledge it is the only other work that uses RST to improve the long form text generation ability of a language model. It has three modules: an RST planner, an encoder and a decoder. These modules are combined to form a VAE structure wherein the RST planner produces a sequence of discrete hidden vectors representing the sequence of RST relations between EDUs in their text. Then at each generation step, sequential hidden vectors are used to guide the text generation process. For fair comparison we do not pass keyphrases to RSTGen.

\noindent\underline{RSTGen Ablations}.
We evaluate ablations of RSTGen to validate the two significance  features of our proposed RSTGen framework. Firstly, we remove the RST-Aware Attentions (R.A.A.), after which we remove RST Nuclearity, then Relation, and finally node positions.

\paragraph{Metrics}

We use two sets of metrics.
For the Argument Generation experiments, we use the GRUEN \citep{zhu-bhat-2020-gruen} set of metrics, which measure four important features of text: Grammaticality, Non-redundancy, Focus and Structure/Coherence. A final metric combines these four features, and has been shown to correlate strongly with human reviewers. For the Story Generation experiments, we use BLEU \citep{papineni-etal-2002-bleu} focusing on the $n$-gram precision; Distinct-$n$ \citep{distinct2016}, which takes the proportion of distinct $n$-grams relative to all generated $n$-grams, thus producing a non-reference-based measure for diversity; the GRUEN Structure and Coherence metric (G-S\&C); and MS-Jaccard \citep{msjaccard2019} which uses the Jaccard Index to measure how similar the distribution of $n$-grams is between two sets of text.

\subsection{Evaluation Results}
This section presents the performance of RSTGen against baseline models and discusses the performance of RSTGen ablations.

\paragraph{Argument Generation}
Table~\ref{tab:arg_gen_performance1} shows that our proposed RSTGen using the DYPLOC Planner performs on par with CTRL in terms of \emph{Grammar} despite using only one tenth parameters of CTRL. It outperforms all the baselines in 
\emph{Structure and Coherence}. 

Using automatically predicted RST trees as input, we observe degraded performance of RSTGen results. We believe that this can be attributed to our Markovian RST sampling methodology of sampling a child node dependent on its direct parent node. This may produce RST Trees with unrealistic structures. Nevertheless, it still outperforms CTRL in \emph{Redundancy} and \emph{Focus}, implying that the combination of RST relations, nuclearity and key phrases provide strong content guidance.

RSTGen without the RST-aware attention (R.A.A.) experiences a drop in performance across all metrics, especially \emph{Grammar} and \emph{Focus}. During the generation of longer sequences, using R.A.A. ensures that the same key phrase or RST information does not influence adjacent elementary discourse units, leading to more diversity and less repetition in generated text. We posit that the removal of R.A.A. exacerbates the reduced performance of RSTGen for longer texts, a trait that was exemplified in Figure~\ref{fig:tree_edit_distance}.

\paragraph{Story Generation}
Table~\ref{tab:StoryGenerationTest} shows that the R.A.A. variants of our model perform on par with the baseline. 
We observe that the removal of the R.A.A. causes a significant drop in performance, specifically Distinct-4, confirming our findings from the Argument Generation. As the RST relations carries information on the semantics of text, we observe that its removal has a significant effect on the similarity based metrics of BLEU-1, Distinct-4 and MS-Jaccard 3.

\begin{table}[!htb]
\centering
\resizebox{\columnwidth}{!}{%
\begin{tabular}{@{}lcccc@{}}
\toprule
Model                & BLEU-1 $\uparrow$ & Distinct-4 $\uparrow$ & G-S\&C $\uparrow$  &  MS-Jaccard 3 $\uparrow$\\ \midrule
DiscoDVT             & \textbf{24.10}         & 84.66         & -0.73  &\textbf{34.76}             \\
RSTGen (no keyphrase)& 24.01         & \textbf{84.85}          & \textbf{-0.71} &34.51            \\
- R.A.A.             & 23.20         & 82.12         & -0.75  & 33.21   \\
- RST relation          &  22.28        & 81.09         & -0.77 &   33.79        \\
- RST nuclearity        &  22.29        & 81.12         & -0.77 &   33.78        \\
- RST positions          & 22.04        & 80.91         & -0.75 &   33.73        \\ \bottomrule
\end{tabular}%
}
\caption{\textbf{Story Generation Evaluation:} Results for the RSTGen and Baselines on the Writing Prompts story generation dataset. RSTGen performs competitively with DiscoDVT.}
\label{tab:StoryGenerationTest}
\end{table}

\section{Conclusion}
We present a novel controlled text generation framework, RSTGen, which uses fine-control over a Rhetorical Structure Theory based context as a means to improve the coherence and cohesion of generated text. We also leverage the structural information presented by RST to propose an rst aware attention scheme, ensuring that the model attends to the correct information during long form generation. Through investigation of RSTGen's open generation text, we showed that our approach can exhibit a high level of intepretable fine-control over syntactic, semantic and structural features of text.

\section{Ethics Statement}


We acknowledge that our proposed model may be susceptible to learning harmful biases present in the dataset. In and of itself this has the potential to harm minorities, marginalised communities and project stigmas present in society. Further, we recognise that our efforts to improve coherence, cohesion and control might be misused to author offensive or fictitious content. Therefore, we advocate for morally correct and responsible practices in the case of real-world application.

\section*{Acknowledgements}
This work was funded by the the UK Engineering and Physical Sciences Research Council (grant no. EP/T017112/1, EP/V048597/1). YH is supported by a Turing AI Fellowship funded by the UK Research and Innovation (grant no. EP/V020579/1).

\bibliographystyle{acl_natbib}
\bibliography{bibliography}

\clearpage

\appendix

\setcounter{table}{0}
\renewcommand{\thetable}{A\arabic{table}}



\section{RST Schemas}

We list in Table \ref{tab:RST_Relations} the RST relations and necleus that our proposed RSTGen framework utlises. We also provide a schema in Table \ref{tab:RST_Nuclearity} explaining the meaning of Nuclearity labels used by RSTGen.

\begin{table*}[hbt!]
\resizebox{2\columnwidth}{!}{%
\begin{tabular}{@{}lll@{}}
\toprule
Relation             & Nuclei        & Satellite \\ \midrule
Attribution          & the effect & the factor which it can be attributed               \\
Background           & text whose understanding is being facilitated               & text whose understanding is being facilitated              \\
Cause                & action/situation & factor which resulted in action/situation's occurrence                \\
Comparison           & a comparison between two or more subjects/objects &    NA           \\
Condition            & action/situation resulting from the occurrence of a conditioning situation &  conditioning situation\\
Contrast             &  one alternate & the other alternative              \\
Elaboration          & basic information & additional information               \\
Enablement           & an action/event enabled by a factor & the factor                \\
Evaluation           & a situation                & an evaluative comment about the situation               \\
Explanation          & a statement      & the supporting statement to explain the statement               \\
Joint                & a list or dis-junction                & NA               \\
Manner-Means         & the action (being) performed & the manner or means by which the action was performed/achieved \\
Topic-Comment        & a statement such as a question, topic or statement & a paired statement such as an answer / topic-comment or response               \\
Summary              &  a statement & a restatement, that is shorter               \\
Temporal             &  a statement with temporal dependence & factor that is depended on \\ 
Topic-Change         &  a shift from this topic A to & a shift to this topic B              \\
Same-Unit            &  Used to link parts of discourse separated by embedded discourse relation              &     NA          \\
Textual-Organization &  used to link parts of discourse separated by embedded discourse relation              &    NA           \\
Null                & Non-classified &  NA        \\ 
\bottomrule
\end{tabular}}
\caption{\textbf{The RST relations Schema: } A schema providing an interpretation for the RST relations our framework utilises. Some of the descriptions are extracted from \citep{doi:10.1177/1461445606064836} and \citep{Carlson2003} }
\label{tab:RST_Relations}
\end{table*}

\begin{table*}[hbt!]
\resizebox{2\columnwidth}{!}{%
\begin{tabular}{@{}lcl@{}}
\hline
Parent Nodes' Nuclearity & Label & Relationship between sibling does                                                       \\ \hline
Nuclei-Nuclei  & NN                    &  The left sibling's sub-text  is equally as important as the right sibling's sub-text. \\
Nuclei-Sattelite  &   NS                  & The left sibling's sub-text is the important part of the parent node's text.       \\
Sattelite-Nuclei& SN  & The right sibling's sub-text is the important part of the parent node's text.      \\
Null            & Null         & Non-classified \\
\bottomrule
\end{tabular}}
\caption{ \textbf{A RST Nuclearity Schema:} A schema providing an interpretation of the RST nuclearity our framework uses.}
\label{tab:RST_Nuclearity}
\end{table*}

\section{Argument and Story Generation Dataset Details}
\label{apdx:Other_Datasets}
For both the argument generation and story generation experiments our datasets originate from the respective paper's of models in the baseline study. We do not make any alterations to the datasets, but instead simply use it as provided. We refer the reader to and \citet{hua2021dyploc} for the argument generation dataset (i.e., the CMV dataset) and \citet{2018_hierarchical_story_generator} for further details regarding the story generation dataset (i.e., the Writing Prompt dataset).

\section{Reproducibility}
\label{apdx:reproducibility}

\paragraph{Code} The code used to train and evaluate our models can be downloaded from \href{https://github.com/Rilwan-A/RSTGen}{https://github.com/Rilwan-A/RSTGen}. As well as code, the github will contain links to the RST annotated versions of the CMV dataset and Writing Prompts dataset. Access to the full RST Annotated Reddit dataset can be gained upon request.

\paragraph{Repositories}
The RSTGen models were extended from pretrained models in Huggingface's Transformers repository \citep{wolf2019huggingfaces}. RSTGen is initialised using the GPT2-base model with approximately 124M parameters. The neural RST Predictor was initalised using the BART-base model. We used Pytorch-Lightning \citep{falcon2019pytorch} for all our training scripts.

\paragraph{Hardware}
For fine-tuning the RSTGen model on the RST Annotated Dateset, we used 2 GeForce RTX 3090 (24GB). For the Argument Generation Tasks and the Story Generation Task, the RSTGen variants and RST Neural Sampler are fine-tuned using 1 GeForce RTX 3090 (24GB). All training was done using mixed Precision (FP16) to improve memory efficiency.

\paragraph{Fine-tuning} For fine-tuning all variations of RSTGen, we used the Adafactor optimiser with the following parameter settings: scale parameter $=$ False, relative step $=$True, warmup init $=$True, learning rate $=$None, weight decay $=$0.01. Due to the high computational expense required, we did not perform extensive hyper-parameter tuning for our RSTGen models.

When fine-tuning on the RST Annotated Dataset, we used an effective batch size of 44 and trained using an Early Stopping rule allowing for at most one epoch to pass with no improvement. The maximum target sequence length is 270 tokens. The maximum RST Tree Size is 36 parent nodes. The maximum Key Phrase sequence size is 64 tokens. These models take approximately 5 epochs to converge which takes approximately 10 hours.

\section{Keyphrase Extraction using TextRank}
\label{apdx:text_rank}

Our second conditioning factor is the phrases that are important to the generated text. This importance is determined by the steps listed below:

\paragraph{Noun Chunks and Named Entities} Given a text $x$, the noun chunks and named entities are extracted to form a set of sub-phrases $[x^{sp}_1,x^{sp}_2, ...,x^{sp}_N ]\in x$.

\paragraph{Graph Formation} Separately, a graph $G$ is formed from text $x$ by extracting all words $w_i$ that have a Part-Of-Speech tag of either `Adjective', `Noun', `Proper Noun' or `VERB'. These words are lemmatised and form nodes $V_i\in G$.

\paragraph{Edge Creation}
A weighted edge $w_{ij}$ between nodes $V_i,V_j, i\ne j$ has a weight of 1 if the distance between the words corresponding to $V_i,V_j$ is less than some threshold $k$. In other TextRank implementations factors such as word length, position and frequency can be used to scale $w_{ij}$.

\paragraph{Node Scoring}
The TextRank score $\mathrm{S}\left(\mathrm{V}_{\mathrm{i}}\right)$ of a node $V_{i}$ is initialised to a default value. Then Page Rank's adapted Eigenvector centrality measure is used to to calculate the importance of each node $V_i$. 
$\mathrm{S}\left(\mathrm{V}_{\mathrm{i}}\right)$ is iteratively updated using the equation below, until convergence is reached:

\begin{equation}\small
\mathrm{S}_{V}\left(\mathrm{V}_{\mathrm{i}}\right)=(1-\mathrm{d})+\left(\mathrm{d} \times \sum_{j \in \mathcal{N}\left(\mathrm{i}\right)} \frac{w_{\mathrm{ji}} \times \mathrm{S}\left(\mathrm{V}_{\mathrm{j}}\right)}{\sum_{\mathrm{k} \in \mathcal{N}\left(\mathrm{j}\right)}w_{\mathrm{jk}}} \right)\nonumber
\label{eqn:page_rank}
\end{equation}
where $\mathrm{d}$ is a damping factor and set to 0.85 as in \cite{mihalcea-tarau-2004-textrank} and $\mathcal{N}(\mathrm{i})$ are the set of indices of the neighbours of $V_i$.

\paragraph{Keyphrase Scoring}
Given the set of node scores $\mathrm{S}\left(\mathrm{V}_{\mathrm{i}}\right)$, the score of a sub-phrase $x^{sp}_n$, with $L$ words, is computed by summing the scores of the words it contains normalised by its length $+1$ to favor longer $n$-grams. This is shown in the equation below: 
\begin{equation}
\mathrm{S_{kp}}(x^{sp}_n)=\frac{\sum_{l=1}^{L}  \mathrm{S}_{V} \left(\mathrm{V}_\mathrm{l} \right) }{\text { L }+1}\nonumber
\label{eqn:subphrase_score}
\end{equation}

Keyphrase candidates are then ranked and redundant candidates filtered out. Two candidate phrases are considered redundant if they have the same lemmatised form.

\section{Argumentation Generation Examples}

In this section, we show some example text generated from RSTGen in comparison to baseline models, where text with poor coherence is underlined. We can observe that CTRL generates coherent text but with a shorten length. Both PAIR and DYPLOC exhibit some inconsistency in their generated text. On the contrary, RSTGen generates much longer text with its inconsistency appearing to be more subtle.

\begin{table*}[htb!]
\centering  
\resizebox{2\columnwidth}{!}{%
\begin{tabular}{p{0.2\linewidth}p{0.8\linewidth}}
\toprule
CTRL & It is important when you're reading old texts like the Bible to remember that it was made to be analyzed, which the human race has been doing for a couple of hundred years by now. If I were to come up to someone and say "How can you believe in a religion that said all types of animals converged to the Middle East to get on a big boat to be saved?" the person would respond "I have faith. I don't need to question what I have been taught."
   \\ \midrule
PAIR & It is important to remember what reading your texts about your Bible and remember that you've made what you analyzed to human race " but remember about your couple of hundred years ago they were born? \myul{I've been taught to think religion, but I've been taught to have types of animals I haven't been in Middle East} and I've never been on boat that a saved person to respond to their faith in the US. My question is, do you really? 
\\ \midrule
DYPLOC & I think it's important to read the Bible as a text, because it's important to remember that the Bible was made up and analyzed by the human race for a couple hundred years. The question is, what is it taught to us? How do we know that a person responds to their faith? \myul{Religion is the type of animal that converges on a big boat and saves the middle east}.
\\ \midrule
RSTGen &  It is a lot more likely to be analyzed by the human race than any other book of text, or even just some couple of important texts that have been written in an attempt to make sense of what we've done for ourselves and our own species. I'm not sure how you can possibly say this without reading all sorts of different perspectives on these topics \myul{as opposed to being made up about something like "the world" which has never really been studied at all - it's only ever happened once.} The whole point of your argument rests on whether people are actually doing anything wrong with their minds (which isn't always true). It doesn't mean they aren't trying to explain things beyond thinking through them... but rather why there was no reason to believe those questions were asked when someone else said `why did god exist?' What if one person had already read such answers? If another person hadn't understood science then wouldn't he simply put themselves into question every single-made assumptions?
\\ \bottomrule
\end{tabular}}
\caption{ \textbf{Argument Generation Example Output:} We highlight parts of the generated text that correspond to poor coherence. In this example we notice at least one inconsistency in the output for PAIR and DYPLOC. However, the inconsistency in the RSTGen appears to be more subtle. }
\label{tab:LongTextArgument}
\end{table*}

\end{document}